\documentclass[10pt,twocolumn,letterpaper]{article}

\usepackage{cvpr}      % To produce the REVIEW version
\usepackage[accsupp]{axessibility}
\usepackage[dvipsnames]{xcolor}

\definecolor{cvprblue}{rgb}{0.21,0.49,0.74}
\usepackage[pagebackref,breaklinks,colorlinks,citecolor=cvprblue]{hyperref}

\def\paperID{0006} % *** Enter the Paper ID here
\def\confName{CVPR}
\def\confYear{2024}

\usepackage{amsfonts}
\usepackage{amsmath}
\usepackage{array}
\usepackage{booktabs}
\usepackage{color}
\usepackage{enumitem}
\usepackage{float}
\usepackage{mathtools}
\usepackage{multirow}
\usepackage{subcaption}
\usepackage{xfrac}

\newcolumntype{L}[1]{>{\raggedright\let\newline\\\arraybackslash\hspace{0pt}}m{#1}}
\newcolumntype{C}[1]{>{\centering\let\newline\\\arraybackslash\hspace{0pt}}m{#1}}
\newcolumntype{R}[1]{>{\raggedleft\let\newline\\\arraybackslash\hspace{0pt}}m{#1}}
\setlist[itemize]{noitemsep, topsep=0pt}
\setlist[enumerate]{noitemsep, topsep=0pt}

\newcommand{\parens}[1]{\left(#1\right)}
\newcommand{\braces}[1]{\left\{#1\right\}}
\newcommand{\bracks}[1]{\left[#1\right]}

\newcommand{\norm}[1]{\left\Vert#1\right\Vert}

\newcommand\numberthis{\addtocounter{equation}{1}\tag{\theequation}}

\title{Speech2UnifiedExpressions: Synchronous Synthesis of Co-Speech Affective Face and Body Expressions from Affordable Inputs}

\author{Uttaran Bhattacharya\thanks{Work partially done while Uttaran was a Ph.D. student at UMD}\\
Adobe Inc.\\
San Jose, CA, USA
\and
Aniket Bera\\
Purdue University\\
West Lafayette, IN, USA
\and
Dinesh Manocha\\
University of Maryland\\
College Park, MD, USA
}

\begin{document}

\twocolumn[{%
\renewcommand\twocolumn[1][]{#1}%
\maketitle
\begin{center}
    \centering
    \captionsetup{type=figure}
    \includegraphics[width=\textwidth]{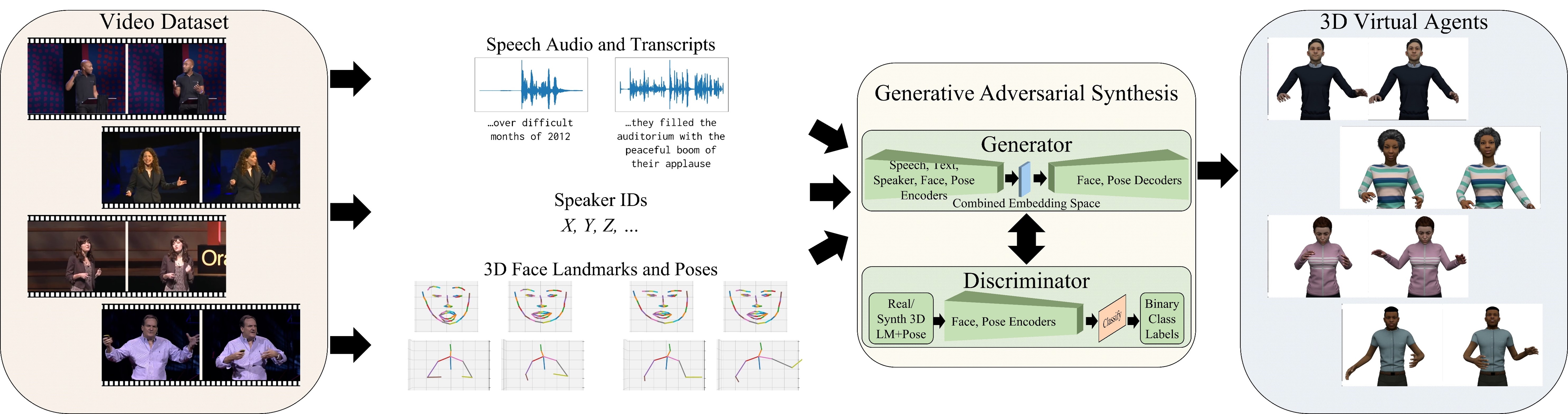}
    \caption{\textbf{Synthesizing unified co-speech 3D face and pose expressions.} Our method uses the speech audio, the corresponding text transcripts, the speaker's unique IDs, and their sparse 3D face landmarks and pose sequences computed from RGB video data. It learns a combined embedding space that captures the correlations between all these inputs and leverages them to generate synchronous affective expressions for faces and poses in a continuous motion space.}
    \label{fig:teaser}
\end{center}%
}]

\begin{abstract}
    We present a multimodal learning-based method to simultaneously synthesize co-speech facial expressions and upper-body gestures for digital characters using RGB video data captured using commodity cameras. Our approach learns from sparse face landmarks and upper-body joints, estimated directly from video data, to generate plausible emotive character motions. Given a speech audio waveform and a token sequence of the speaker's face landmark motion and body-joint motion computed from a video, our method synthesizes the motion sequences for the speaker's face landmarks and body joints to match the content and the affect of the speech. We design a generator consisting of a set of encoders to transform all the inputs into a multimodal embedding space capturing their correlations, followed by a pair of decoders to synthesize the desired face and pose motions. To enhance the plausibility of synthesis, we use an adversarial discriminator that learns to differentiate between the face and pose motions computed from the original videos and our synthesized motions based on their affective expressions. To evaluate our approach, we extend the TED Gesture Dataset to include view-normalized, co-speech face landmarks in addition to body gestures. We demonstrate the performance of our method through thorough quantitative and qualitative experiments on multiple evaluation metrics and via a user study. We observe that our method results in low reconstruction error and produces synthesized samples with diverse facial expressions and body gestures for digital characters. The relevant source code and dataset are available at \url{https://github.com/UttaranB127/speech2unified_expressions}.
\end{abstract}

% --------------------------------------------%

\section{Introduction}\label{sec:intro}
Human communications through digital platforms and virtual spaces are prevalent in many applications, including online learning~\cite{online_learning1,online_learning2,online_learning3}, virtual interviewing~\cite{interviewing}, counseling~\cite{simsensei}, social robotics~\cite{cospeech_gestures}, automated character designing~\cite{game_development}, storyboard visualizing for consumer media~\cite{gesticulator,script_visualization}, and creating large-scale metaverse worlds~\cite{omniverse}. Simulating immersive experiences in such digital applications necessitates the development of plausible human avatars with expressive faces and body motions. This is a challenging problem to approach at scale, given the necessity and diversity of human expressions in human-human interactions~\cite{emotions_in_social_relations,emotions_in_context}.
Further, humans express simultaneously through multiple cues or \textit{modalities}, such as their speech, facial expressions, and body gestures~\cite{m3er}, increasing the dimensionality of the problem. The emotional expressions from these different modalities are also synchronous, \textit{i.e.}, they follow the same rhythm of communication and complement each other to convey a sense of presence~\cite{mastery_of_movement}. 

In this paper, we consider the problem of synthesizing 3D digital human motions with synchronous facial expressions and upper-body gestures aligned with speech audio inputs. Given the speech audio, existing approaches commonly tackle the sub-problems of ``talking heads''~\cite{audio_driven_face} -- synthesizing lip movements and facial expressions given the speech audio, and co-speech gesture synthesis~\cite{trimodal} -- synthesizing poses for upper-body gestures, including head motions. Recent approaches synthesize head and body motions simultaneously~\cite{speech_driven_conv_gestures,talkshow}, but consider a limited set of speakers and their expressions. More general motion synthesis methods consider full-body motions from various modalities, including text prompts~\cite{compositional_animations,mofusion}, object interactions~\cite{imos,grip}, and guidance motions~\cite{priormdm,remos}, but do not combine modalities (such as face and pose) in the output. The inherent difficulty in synthesizing expressions synchronized across diverse speakers is to under the correlations between the modalities for both the expressions and the individual styles~\cite{face_and_gesture_emotions}. In other words, not only is the combined space of the multimodal expressions very high-dimensional, but only a small fraction of that space corresponds to \textit{valid} expressions for different speakers. Moreover, existing approaches generally require specialized data, such as dense 3D face scans~\cite{voca} and motion-captured gestures~\cite{iemocap,t2g}, often employ parameter-dense and compute-heavy approaches, such as those based on denoising diffusion~\cite{motiondiffuse,gesturediffuclip} to provide meaningful results. By contrast, we aim to develop a lightweight method for synchronous co-speech face and pose expressions by leveraging large-scale video datasets~\cite{cospeech_gestures}, paving the way to synthesize fully expressive 3D digital humans for democratized use.

\noindent\textbf{Main Contributions.} We present a multimodal learning method to synthesize 3D digital characters with synchronous affective expressions on faces and upper-body poses, given speech audio. We also consider both intra- and inter-speaker variabilities by random sampling on a latent space for speakers.
Our main contributions include:

\begin{itemize}
    \item \textbf{Synchronous co-speech face and pose expression synthesis.} Our method simultaneously synthesizes face and upper-body pose expressions given speech audio through a generative multimodal embedding space and an affective discriminator.
    Our method reduces the mean absolute errors on the face landmarks by $30\%$, and the body poses by $21\%$, compared to the respective baselines for faces and poses, thereby indicating measurable benefits over asynchronously combining the synthesized outputs of the two modalities.

    \item \textbf{Using data from affordable commodity cameras.} In contrast to facial expression synthesis using dense 3D face scans or gesture synthesis from expensive motion-captured data, our method only relies on sparse face landmarks and pose joints obtainable from commodity hardware such as video cameras. As a result, our method scales affordably to large datasets and is applicable in large-scale social applications.

    \item \textbf{Plausible motions, evaluation metric for facial expressions.} Through quantitative evaluations and user studies, we verify that our synthesized synchronous expressions are satisfactory to human observers. We also propose the Fr\'echet Landmark Distance to evaluate the quality of the synthesized face landmarks.

    \item \textbf{TED Gesture+Face Dataset.} We extend the TED Gesture Dataset to include 3D face landmarks extracted from the raw videos that we denoise and align with the poses. We release this multimodal dataset of speech audio, 3D face landmarks, and 3D body pose joints with our paper and the associated source code.
\end{itemize}

\begin{figure*}
    \centering
    \includegraphics[width=\textwidth]{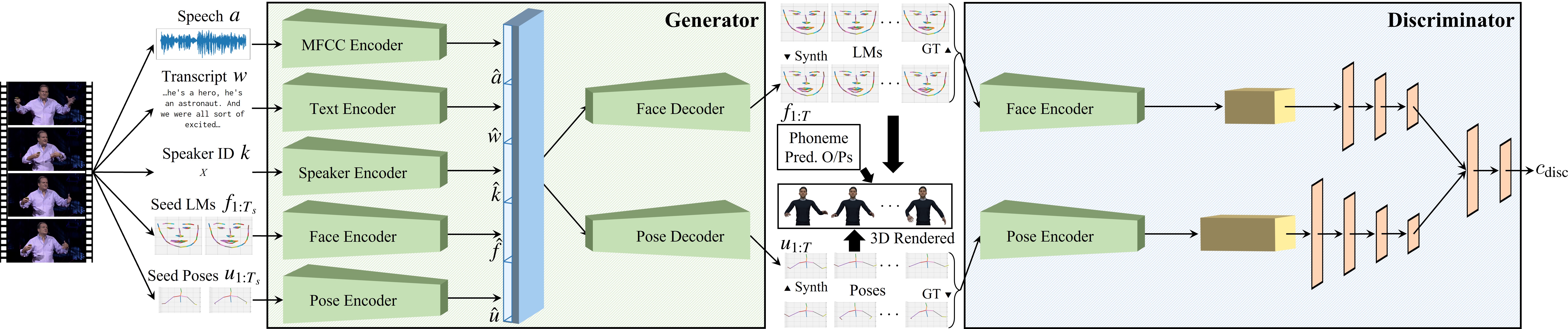}
    \caption{\textbf{Network architecture for synchronous synthesis of co-speech face and pose expressions.} Our generator encodes all the inputs: the speech audio, the corresponding test transcript, the speaker ID, the seed 3D face landmarks, and the seed 3D poses into a multimodal embedding space. It decodes variables from this space to produce the synchronized sequences of co-speech 3D face landmarks and poses. Our discriminator classifies these synthesized sequences and the corresponding ground truths (3D motions of the original speakers), computed directly from the videos, into two different classes based both on their plausibility and their synchronous expressions. To obtain our rendered 3D character motions, we combine the outputs of our generator with our phoneme predictor network and map them to 3D meshes.}
    \label{fig:s2ue_network}
\end{figure*}

% --------------------------------------------%

\section{Related Work}
We briefly review prior work on perceiving multimodal affective expressions, particularly from faces, speech, and gestures, and synthesis of co-speech face and pose motions.

\paragraph{Perceiving Multimodal Affective Expressions.}
Studies in psychology and affective computing indicate that humans express emotions simultaneously through multiple modalities, including facial expressions, prosody and intonations of the voice, and body gestures~\cite{mer1,m3er}. Methods for detecting facial expressions~\cite{fer1} generally depend on facial action units~\cite{mtcnn}. Methods for detecting various affective vocal patterns commonly use Mel-Frequency Cepstral Coefficients (MFCCs)~\cite{mfcc_ser}. Methods to detect emotions from body gestures use physiological features, such as arm swings, spine posture, and head motions that are either pre-defined~\cite{step,gzsl_gestures} or learned automatically from the gestures~\cite{taew}. The emotions themselves can be represented either as discrete categories, such as the Ekman emotions~\cite{ekman_emotions} or as combinations of continuous dimensions, such as the Valence-Arousal-Dominance (VAD) model~\cite{vad}. In
our work, we leverage the current approaches for detecting facial, vocal, and pose expressions to design our co-speech face and gesture synthesis method. While we do not explicitly consider specific emotions, our representation implicitly considers emotions in the continuous VAD space, leading to appropriately expressive face and pose synthesis.

\paragraph{Synthesizing Co-Speech Motions.}
We consider digital characters with faces and body gestures.

\noindent{\textit{Co-Speech Facial Expressions.}}
\citeauthor{hmm_talking_heads}~\cite{hmm_talking_heads} compute controllable parameters for synthesizing talking heads with desired facial expressions using a Hidden Markov Model and MFCCs of the speech audio. Recent techniques automate the facial motions for large-scale synthesis using generative paradigms, such as VAEs~\cite{head_motion_ae} and GANs~\cite{head_gan}. \citeauthor{audio_driven_face}~\cite{audio_driven_face} train a DNN to map speech audio to 3D face vertices conditioned on learned latent features corresponding to different facial expressions. \citeauthor{visemenet}~\cite{visemenet}, learn sequences of predefined visemes using LSTM networks from audio. \citeauthor{voca}~\cite{voca} propose a dataset of 4D face scans and learn per-vertex offsets to synthesize the face motions from audio. \citeauthor{meshtalk}~\cite{meshtalk} learn co-speech facial motions using dense face meshes by disentangling facial features correlated and uncorrelated with speech. \citeauthor{emotional_talking_head}~\cite{emotional_talking_head} focus on adding emotional expressions to the faces. \citeauthor{lipsync}~\cite{lipsync} focus on the accuracy of the lip movements and use an autoregressive approach to synthesize 3D vertex sequences for the lips synced with the speech audio. In contrast to these approaches, our facial expression synthesis method uses much sparser 3D face landmarks detected from real-world videos with arbitrary orientations and lighting conditions of the faces w.r.t. the cameras, and synthesizes mutually coherent facial and pose expressions.

\noindent{\textit{Co-Speech Gestures.}}
Co-speech gesture synthesis is a special case of gesture stylization, where the style refers to the pose expressions inferred from and aligned with the speech. This line of work has been richly explored~\cite{3d_pose_spatial_temporal_estimation,speech_templates,a2g,rhythmic_gesticulator,motion_matching,a2g_video,disco,h2ag}. \citeauthor{individual_gesture_styles}~\cite{individual_gesture_styles} propose a method to synthesize speaker-specific co-speech gestures by training a neural network given their identities and individual gesticulation patterns. \citeauthor{multi_adversarial_gestures}~\cite{multi_adversarial_gestures} additionally propose using adversarial losses in the training process to improve the fidelity of the synthesized gestures. \citeauthor{trimodal}~\cite{trimodal} extend the concept of individualized gestures to a continuous space of speakers to incorporate natural variability in the synthesized gestures even for the same speaker. \citeauthor{s2ag}~\cite{s2ag} build on top of~\cite{trimodal} to improve the affective expressions in the co-speech gestures. More recent methods have also explored diffusion-based approaches for editability~\cite{gesturediffuclip}. Our method conditions the gesture synthesis on both the input speech and the synthesized facial expressions.

\noindent{\textit{Co-Speech Multimodal Expressions.}}
Co-speech face and upper-body generation has gained particular interest recently, primarily due to the availability of rich 3D datasets of popular speakers~\cite{speech_driven_conv_gestures}. Current approaches train adversarial encoder-decoder models on datasets of one speaker at a time~\cite{speech_driven_conv_gestures} and use vector quantization for tokenized generation using a transformer~\cite{talkshow}. These approaches consider a fixed set of speakers and lose fine-grained expressions when using quantization. In our work, we consider the combined continuous space of affective face and body expressions and develop a network generalizable to multiple speakers.

% --------------------------------------------%

\section{Synchronous Face and Pose Synthesis}
Given a speech audio waveform $a$, the corresponding text transcript $w$, the speaker's unique ID $k$ in a set of speakers $K$, and the associated seed face landmark deltas $f_{1:T_s}$ and seed pose unit vectors $u_{1:T_s}$, $T_s$ being the number of seed time steps, we synthesize the synchronous sequences of face landmark deltas $f_{1:T}$ and pose unit vectors $u_{1:T}$ for the speaker for the $T$ prediction time steps ($T \gg T_s$), matching the content and the affect in their speech. We describe our end-to-end pipeline, including a detailed description of our inputs and outputs and their usage. We also provide the details of obtaining these facial landmarks and poses from input videos.

\subsection{Face and Pose Preprocessing from Video}~\label{subsec:data_preprocessing}
Given a video, we use Multi-Task Cascaded CNNs~\cite{mtcnn} to extract the 3D face landmarks. Since the faces can be arbitrarily oriented w.r.t. the camera, we rigidly transform the face landmarks per frame to a reference frame in the normalized view, where the face looks towards the camera. For each frame in the input video, we use the rotation and the translation given by the Umeyama method~\cite{umeyama} to map the face landmarks in that frame to the face landmarks in the reference frame. We also use similarly view-normalized 3D poses. View normalization is helpful for two key reasons. First, it eliminates relative camera movements across the video frames and prevents a learning-based method from confusing camera movements with face and pose expression changes. Second, a frontal view offers maximal visibility of the faces and the poses, and minimizes errors in detecting the 3D face landmarks and body joints.

\subsection{Computing Face and Pose Expressions}~\label{subsec:face_and_poses}
We consider a reference neutral expression $\mathcal{F} \in \mathbb{R}^{L \times 3}$ for each user, $L$ being the number of face landmarks. To synthesize facial expressions, we compute the relative motion of each landmark w.r.t. the reference expression. Specifically, we obtain the configuration $\mathcal{F}_t$ at time step $t$ as
\begin{equation}
    \mathcal{F}_t = \mathcal{F} + f_t,
\end{equation}
where $f_t \in \mathbb{R}^{L \time 3}$ denotes the set of relative motions of the landmarks w.r.t. $\mathcal{F}$ at time step $t$.

\begin{figure*}[t]
    \centering
    \begin{subfigure}[b]{0.24\textwidth}
        \centering
        \includegraphics[width=0.9\textwidth]{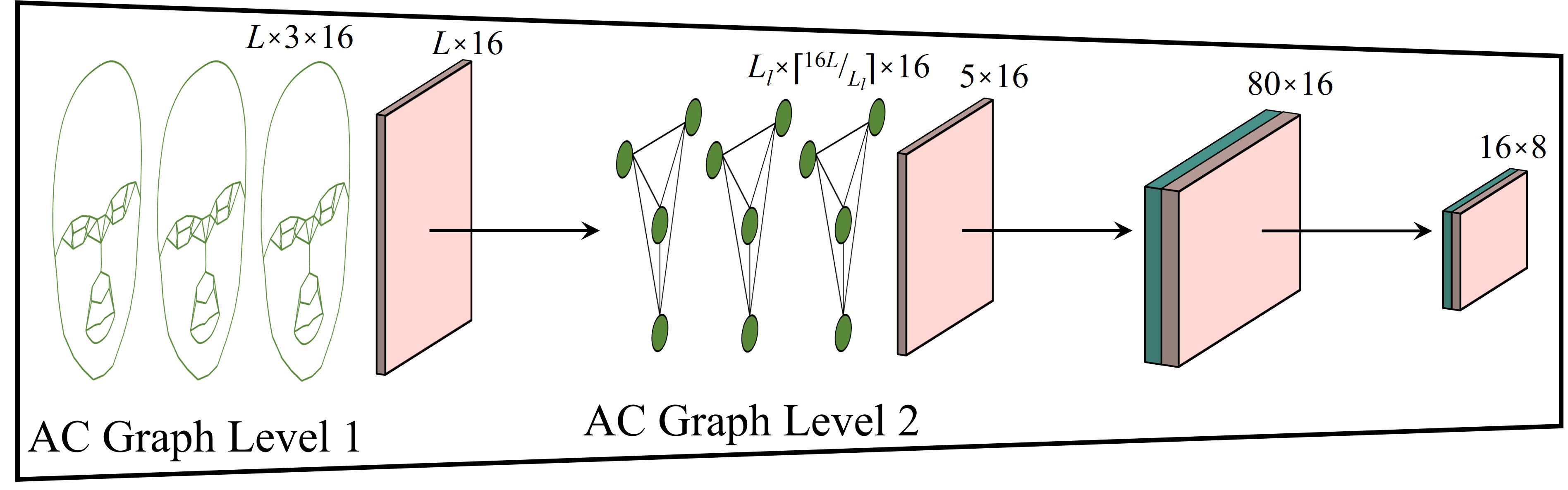}
        \caption{Face encoder}
        \label{fig:face_encoder}
    \end{subfigure}
    \hfill
    \begin{subfigure}[b]{0.24\textwidth}
        \centering
        \includegraphics[width=0.9\textwidth]{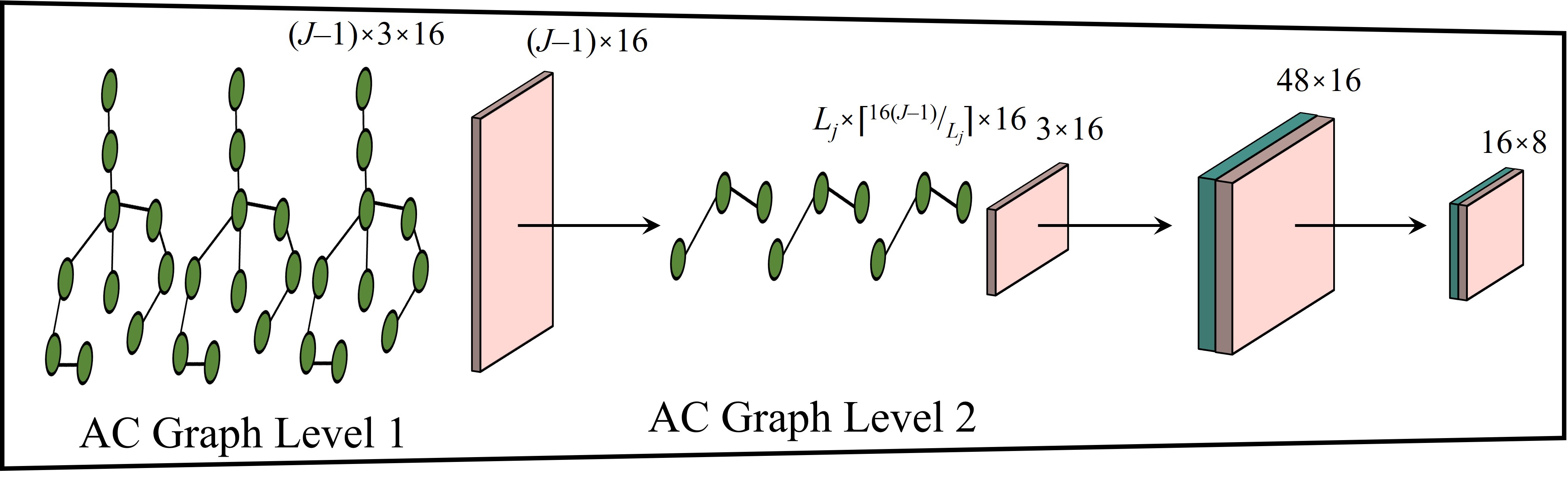}
        \caption{Pose Encoder}
        \label{fig:pose_encoder}
    \end{subfigure}
    \hfill
    \begin{subfigure}[b]{0.24\textwidth}
        \centering
        \includegraphics[width=0.9\textwidth]{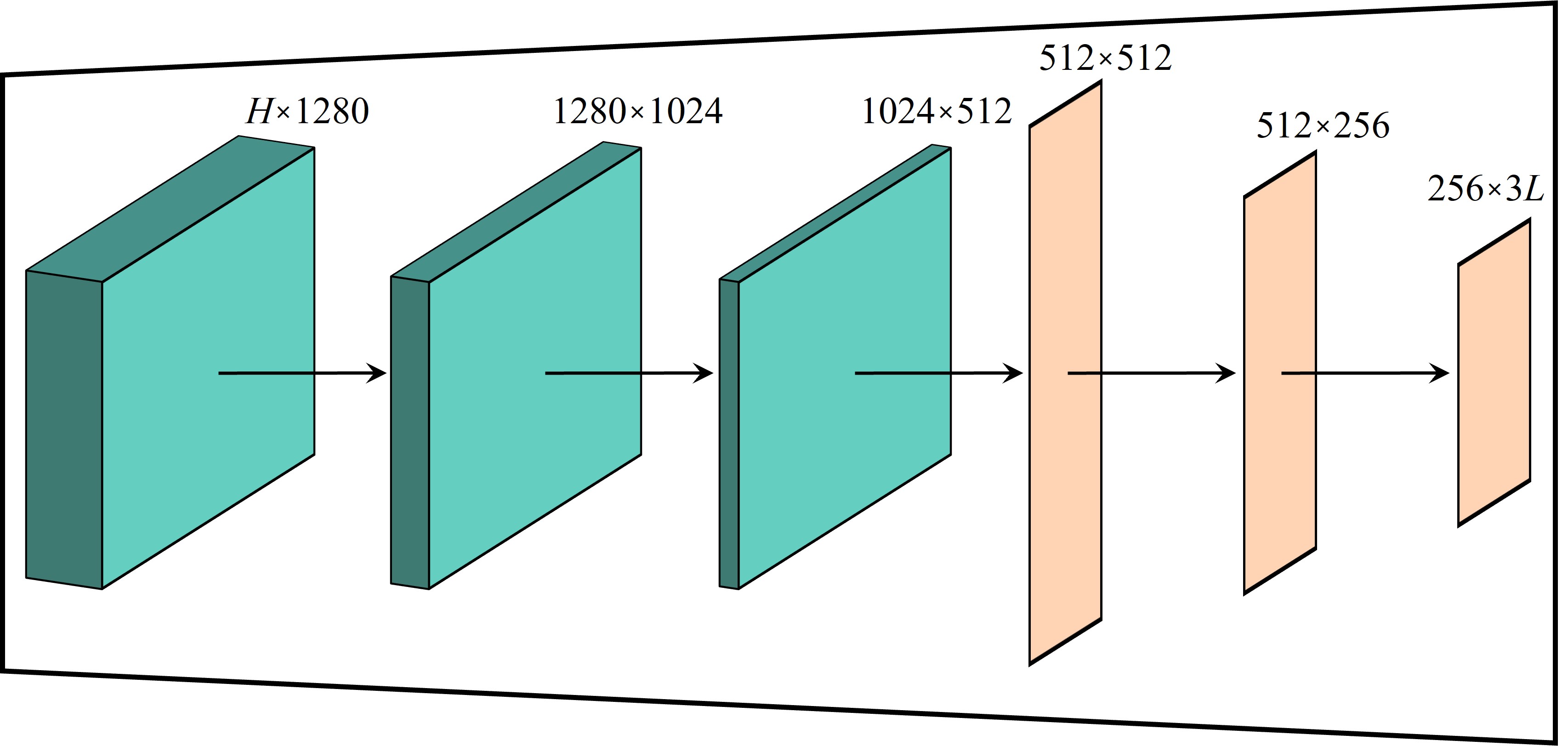}
        \caption{Face Decoder}
        \label{fig:face_decoder}
    \end{subfigure}
    \hfill
    \begin{subfigure}[b]{0.24\textwidth}
        \centering
        \includegraphics[width=0.9\textwidth]{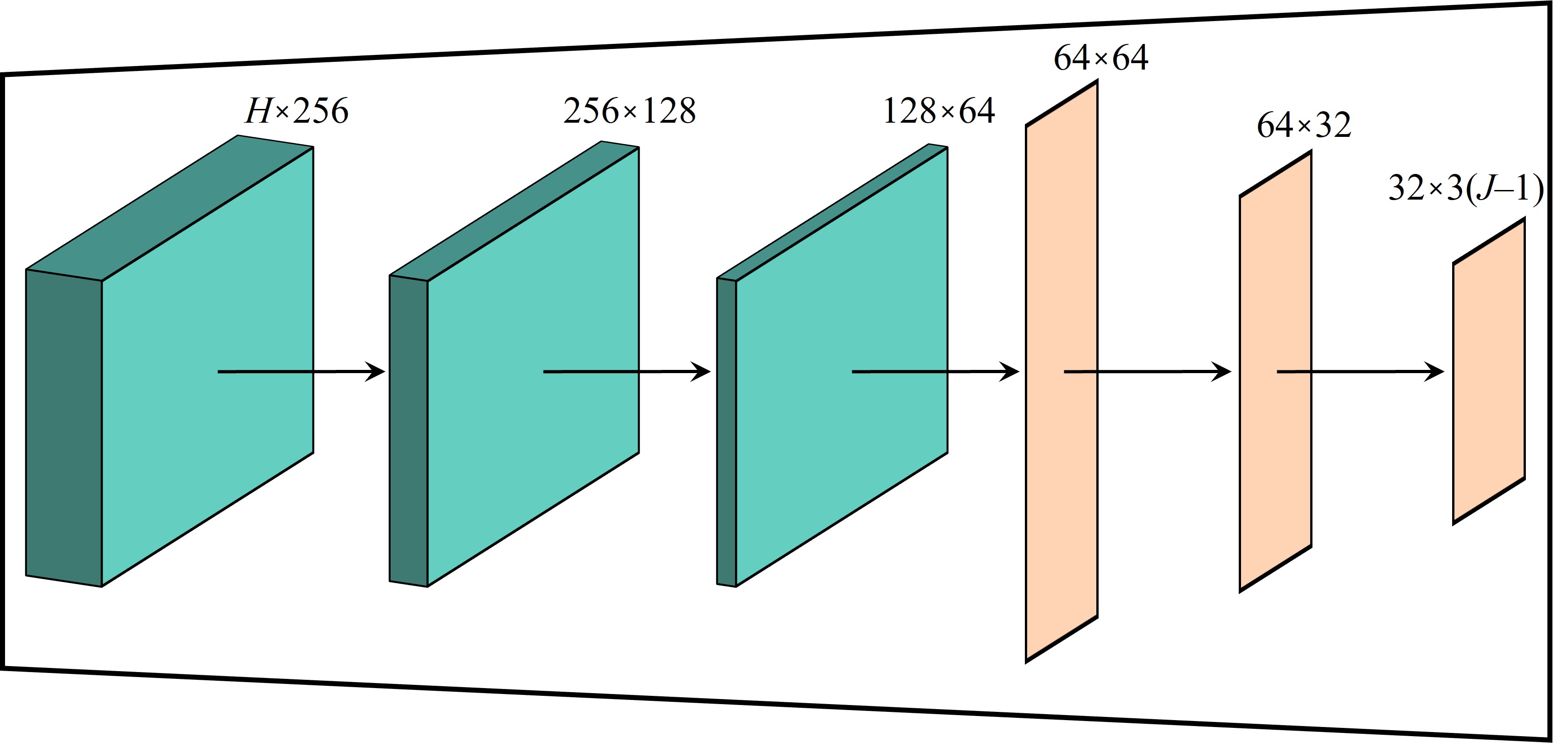}
        \caption{Pose Decoder}
        \label{fig:pose_decoder}
    \end{subfigure}
    \caption{\textbf{Face and pose encoders and decoders.} We show their architectures with the layer sizes denoted (details in Sec.~\ref{subsubsec:enc_aff_exp}). Our architectures depend on the hierarchical anatomical component (AC) graphs for both faces and poses that efficiently learn their corresponding affect representations using spatial-temporal graph convolutions (green nodes and edges), 2D convolutions (teal blocks), 2D batch normalizations (pink blocks), and fully-connected layers (orange planes).}
    \label{fig:face_and_pose_encoder_and_decoder}
\end{figure*}

On the other hand, we assume the body joints are rigidly connected by the bones. We represent each user's body joints as 3D point vectors $\mathcal{P} \in \mathbb{R}^{J \times 3}$ in a global coordinate space, where $J$ is the number of joints. We consider directed line vectors connecting adjacent joints. The direction is along the path from the root (pelvis) joint to the end effectors (such as wrists). These 3D point and line vectors collectively form a directed tree with $J$ nodes and $J-1$ edges. We assume that the magnitudes of these line vectors correspond to the bone lengths and that these magnitudes are known and fixed. To synthesize the users' body gestures, we compute the orientations of these line vectors at each time step $t$ in the reference frame of the global coordinate space. Specifically, for each bone $b$ with bone length (magnitude) $\norm{b}$ and connecting the source joint $s_b\parens{t}$ to the destination joint $d_b\parens{t}$ at time step $t$, we compute a unit vector $u_t$ such that
\begin{equation}
    d_b = s_b + \frac{\norm{b}}{\norm{u_t}}u_t.
\end{equation}
We do not assume any locomotion, \textit{i.e.}, we consider the root joint is fixed at the global origin at all the time steps.

\subsection{Synthesizing Faces and Poses}~\label{subsec:face_pose_synthesis}
Our network architecture (Fig.~\ref{fig:s2ue_network}) consists of a phoneme predictor to predict the lip shapes corresponding to the audio and a generator-discriminator pair to synthesize plausible co-speech face and pose expressions. We design our phoneme predictor following prior approaches~\cite{lipsync} and provide its details in Sec.~\ref{subsubsec:phoneme_pred}. Our generator follows a multimodal learning strategy. It consists of separate encoders to transform the speech audio, the text transcript, the speaker ID, the seed face landmark deltas, and the seed pose unit vectors into a latent embedding space representing their correlations. It subsequently synthesizes the appropriate face and pose motions from this multimodal embedding space. Our discriminator enforces our generator to synthesize plausible face and pose motions in terms of their affective expressions. To this end, we use the same encoder architecture for the faces and the poses as in our generator, but learned separately. We describe each of the components of our generator and discriminator.

\subsubsection{Encoding Speech, Text, and Speaker IDs}
We use the Mel-Frequency Cepstral Coefficients (MFCCs) for the speech audio to accurately capture the affective intonations in the speech, and use an MFCC encoder to obtain speech-based latent embeddings $\hat{a} \in \mathbb{R}^{T \times D_a}$ of dimension $D_a$ as
\begin{equation}
    \hat{a} = \textrm{MFCCEncoder}\parens{a; \theta_{\textrm{MFCC}}},
\end{equation}
where $\theta_{\textrm{MFCC}}$ represents the trainable parameters.

Similarly, we use the sentiment-aware FastText~\cite{fasttext_sentiment_analysis} embeddings of the words in the transcript and a convolution-based text encoder to obtain the text-based latent embeddings $\hat{w} \in \mathbb{R}^{T \times D_w}$ of dimensions $D_w$ as
\begin{equation}
    \hat{w} = \textrm{TextEncoder}\parens{w; \theta_{\textrm{text}}},
\end{equation}
where $\theta_{\textrm{text}}$ represents the trainable parameters.

We also represent the speaker IDs $k \in \braces{0, 1}^K$ as one-hot vectors for a total of $K$ speakers and use a speaker encoder to obtain the parameters $\mu_k \in \mathbb{R}^{D_k}$ and $\Sigma_k \in \mathbb{R}_{+}^{D_k \times D_k}$ of a latent distribution space of dimension $D_k$ as
\begin{equation}
    \mu_k, \Sigma_k = \textrm{SpeakerEncoder}\parens{k; \theta_{\textrm{speaker}}},
\end{equation}
where $\theta_{\textrm{speaker}}$ represents the trainable parameters. The latent distribution space enables us to sample a random vector $\hat{k}$ representing a speaker who is an arbitrary combination of the $K$ speakers in the dataset. This allows for variations in the synthesized motions even for the same original speaker by slightly perturbing their speaker IDs in the latent distribution space, leading to more plausible results on multiple runs of our network. To learn faces and poses with appropriate expressions, we represent them as multi-scale graphs and encode them using graph convolutional networks.

\subsubsection{Encoding Affective Expressions}\label{subsubsec:enc_aff_exp}
Our face landmarks are based on action units~\cite{mtcnn}.
We represent the sequence of 3D landmarks $f_{1:T_s} \in \mathbb{R}^{T_s \times L \times 3}$ as a spatial-temporal anatomical component (AC) graph. Spatially, we consider landmarks belonging to the same anatomical component (Sec.~\ref{subsec:face_and_poses}) and nearest landmarks across different anatomical components to be adjacent. Temporally, all landmarks are adjacent to their temporal counterparts (same nodes at different time steps) within a predetermined time window. We consider the eyes, the nose, the lips, and the lower jaw as the anatomical components. We show the face landmarks graph in Fig.~\ref{fig:face_encoder} with all the intra- and inter-anatomical-component adjacencies marked with lines. We apply a sequence of spatial-temporal graph convolutions on this graph to learn from the localized motions of the landmarks and obtain embeddings $\tilde{f} \in \mathbb{R}^{T_s \times L \times D_f}$ of feature dimension $D_f$ as
\begin{equation}
    \tilde{f} = \textrm{STGCN}_{f}\parens{f_{1:T_s}; \theta_{\textrm{STGCN}_{f}}},
    \label{eq:stgcn_face_1}
\end{equation}
where $\theta_{\textrm{STGCN}_f}$ represents the trainable parameters. We obtain a face anatomy graph from the landmarks graph, where we consider the nodes to represent entire anatomical components and the graph to be fully connected. To compute such a graph, we append the features of intra-anatomical-component nodes in the graph into collated features $l \in \mathbb{R}^{T_s \times L_l \times n_lD_f}$, where $L_l$ denotes the number of anatomical components and $n_l$ denotes the number of landmark nodes within each anatomical component. We take $n_l$ as the number of nodes in the anatomical component with the most landmarks and perform zero padding as appropriate to obtain the full collated features for the other components. This hierarchically pooled representation provides a ``higher-level'' view of the face and helps our network learn from the correlations between the motions of the different anatomical components. Specifically, we use another set of spatial-temporal graph convolutions to obtain the embeddings $\tilde{l} \in \mathbb{R}^{T_s \times L_l \times D_l}$ of feature dimension $D_l$ as
\begin{equation}
    \tilde{l} = \textrm{STGCN}_{l}\parens{l; \theta_{\textrm{STGCN}_{l}}},
    \label{eq:stgcn_face_2}
\end{equation}
where $\theta_{\textrm{STGCN}_l}$ represents the trainable parameters. Collectively, the landmarks graph and the face anatomy graph provide complementary information to our network to encode and synthesize the required facial expressions at both the macro (anatomy) and the micro (landmark) levels. To complete our encoding, we flatten out the features of all the anatomical components in $\tilde{l}$, \textit{i.e.}, reshaping such that $\tilde{l} \in \mathbb{R}^{T_s \times L_lD_l}$, and transform them using standard convolutional layers on the flattened feature channel and the temporal channel separately. This gives us our latent space embeddings $\hat{l} \in \mathbb{R}^{T \times D_{\tilde{l}}}$ as
\begin{equation}
    \hat{l} = \textrm{ConvT}_{\tilde{l}}\parens{\textrm{ConvS}_{\tilde{l}}\parens{\tilde{l}; \theta_{\textrm{ConvS}_{\tilde{l}}}}; \theta_{\textrm{ConvT}_{\tilde{l}}}},
\end{equation}
where $\theta_{\textrm{ConvS}_{\tilde{l}}}$ and $\theta_{\textrm{ConvT}_{\tilde{l}}}$ represent the trainable parameters.

For the pose representation, we consider a pose graph of the upper body with $J-1$ bones represented with line vectors $u_{1:T_s}$ (Fig.~\ref{fig:pose_encoder}). We consider bones connected to each other or connected through a third bone to be adjacent. We use a set of spatial-temporal graph convolutions to leverage the localized motions of these bones and obtain embeddings $\tilde{u} \in \mathbb{R}^{T_s \times D_u}$ of feature dimension $D_u$ as
\begin{equation}
    \tilde{u} = \textrm{STGCN}_{u}\parens{u_{1:T_s}; \theta_{\textrm{STGCN}_{u}}},
    \label{eq:stgcn_pose_1}
\end{equation}
where $\theta_{\textrm{STGCN}_u}$ represents the trainable parameters. Similar to the face landmarks, we also consider a hierarchically pooled representation of the bones $v \in \mathbb{R}^{T_s \times L_j \times n_jD_u}$, where $L_j=3$ are the three anatomical components, the torso and the two arms, represented as single nodes each consisting of $n_j$ nodes from the pose graph. In the pose anatomy graph, we consider the two arms to be adjacent to the torso but not to each other, as they can move independently. We apply a second set of spatial-temporal graph convolutions on the collated features $v$ to obtain the embeddings $\tilde{v} \in \mathbb{R}^{T_s \times L_j \times D_v}$ as
\begin{equation}
    \tilde{v} = \textrm{STGCN}_{v}\parens{v; \theta_{\textrm{STGCN}_{v}}}
    \label{eq:stgcn_pose_2}
\end{equation}
where $\theta_{\textrm{STGCN}_v}$ represents the trainable parameters. To subsequently obtain the latent space embeddings $\hat{v} \in \mathbb{R}^{T \times D_{\tilde{v}}}$, we apply separate spatial and temporal convolutions on the flattened graph-convolved features $\tilde{v} \in \mathbb{R}^{T_s \times L_jD_v}$, as
\begin{equation}
    \hat{v} = \textrm{ConvT}_{\tilde{v}}\parens{\textrm{ConvS}_{\tilde{v}}\parens{\tilde{v}; \theta_{\textrm{ConvS}_{\tilde{v}}}}; \theta_{\textrm{ConvT}_{\tilde{v}}}},
\end{equation}
where $\theta_{\textrm{ConvS}_{\tilde{v}}}$ and $\theta_{\textrm{ConvT}_{\tilde{v}}}$ represent the trainable params.

\begin{figure}[t]
    \centering
    \includegraphics[width=\columnwidth]{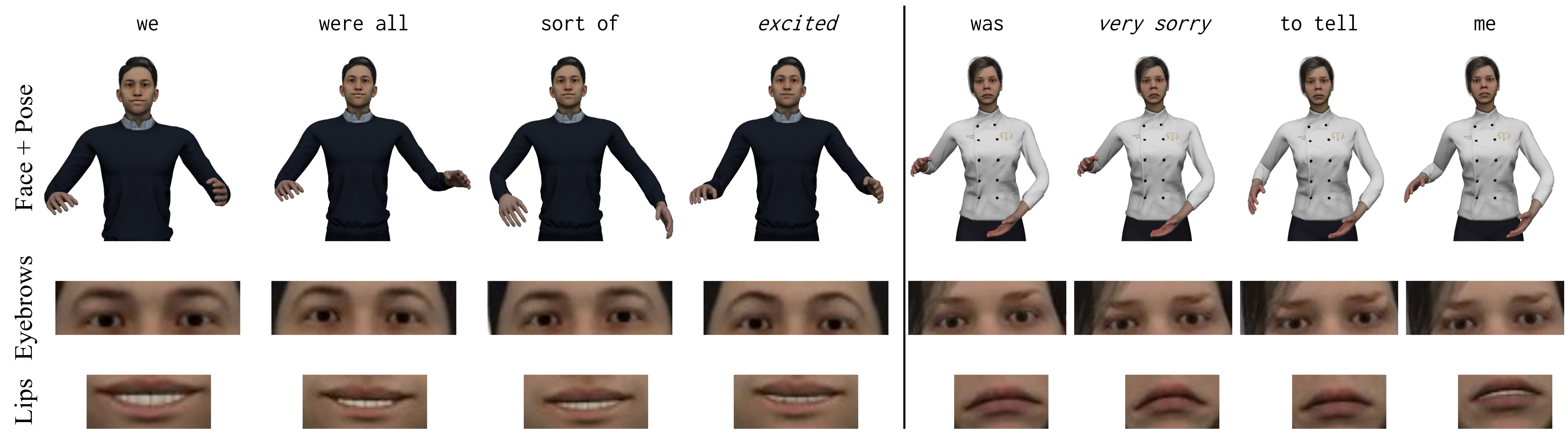}
    \caption{\textbf{Qualitative results.} Snapshots from two of our synthesized samples showing the text transcript of the speech and the corresponding face and pose expressions (row 1). We also zoom in on the eyebrow (row 2) and lip (row 3) expressions for better visualization. We observe a smile, raised eyebrows, and stretched arms (left) for the word `excited', and frowns on the eyebrows and lips (right) for the words `very sorry'.}
    \label{fig:qualitative_results}
\end{figure}

\subsubsection{Synthesizing Synchronous Motions}
Our synchronous synthesis relies on learning the multimodal distributions of the individual modalities of audio, text, speaker ID, face expressions, and pose expressions, given their individual distributions. To this end, we append all the latent space embeddings --- $\hat{a}$ for the audio, $\hat{w}$ for the text, $\hat{k}$ for the random speaker representation, repeated over all the $T$ time steps, $\hat{l}$ for the seed landmarks and $\hat{v}$ for the seed poses --- into a vector $\hat{e} \in \mathbb{R}^{T \times H}$ representing a multimodal embedding space of all the inputs. Here, $H = D_a + D_w + D_k + D_{\tilde{l}} + D_{\tilde{v}}$ denotes the latent space dimension. On training, our network learns the correlations between the different inputs in this multimodal embedding space. To synthesize our face landmark motions $f_{1:T} \in \mathbb{R}^{T \times L \times 3}$, we apply separate spatial and temporal convolutions on the multimodal embeddings $\hat{e}$ to capture localized dependencies between the feature values followed by fully-connected layers capturing all the dependencies between the feature values (Fig.~\ref{fig:face_decoder}), as
\begin{equation}
    \resizebox{0.88\columnwidth}{!}{
        $f_{1:T} = \textrm{FC}_{f\hat{e}}\parens{\textrm{ConvS}_{f\hat{e}}\parens{\textrm{ConvT}_{f\hat{e}}\parens{\hat{e}; \theta_{\textrm{ConvT}_{f\hat{e}}}}; \theta_{\textrm{ConvS}_{f\hat{e}}}}; \theta_{\textrm{FC}_{f\hat{e}}}}$,
    }
\end{equation}
where $\theta_{\textrm{ConvT}_{f\hat{e}}}$, $\theta_{\textrm{ConvS}_{f\hat{e}}}$, and $\theta_{\textrm{FC}_{f\hat{e}}}$ represent the trainable parameters. The output $f_{1:T}$ from the fully connected layers has shape $T \times 3L$, which we reshape into $T \times L \times 3$ to get our desired 3D face landmark sequences.

We similarly synthesize the line vectors $u_{1:T} \in \mathbb{R}^{T \times \parens{J-1} \times 3}$ using separate spatial and temporal convolutions on the multimodal embeddings $\hat{e}$, followed by fully-connected layers (Fig.~\ref{fig:pose_decoder}), as
\begin{equation}
    \resizebox{0.88\columnwidth}{!}{
        $u_{1:T} = \textrm{FC}_{u\hat{e}}\parens{\textrm{ConvS}_{u\hat{e}}\parens{\textrm{ConvT}_{u\hat{e}}\parens{\hat{e}; \theta_{\textrm{ConvT}_{u\hat{e}}}}; \theta_{\textrm{ConvS}_{u\hat{e}}}}; \theta_{\textrm{FC}_{u\hat{e}}}}$,
    }
\end{equation}
where $\theta_{\textrm{ConvT}_{u\hat{e}}}$, $\theta_{\textrm{ConvS}_{u\hat{e}}}$, and $\theta_{\textrm{FC}_{u\hat{e}}}$ represent the trainable parameters. Given the synthesized face and pose motions, we use our discriminator to determine how well their affective expressions match the corresponding ground truths in the training data. We obtain our ground truths as the 3D face landmarks and the 3D pose sequences computed from the full training video data.

\subsubsection{Determining Plausibility Using Discriminator}
Our discriminator takes in the synchronously synthesized face motions $f_{1:T}$ and pose motions $u_{1:T}$, and encodes them using encoders with the same architecture as our generator (Sec.~\ref{subsubsec:enc_aff_exp}), with only the number of input time steps being $T$ instead of $T_s$. This gives us the corresponding latent space embeddings $\hat{l}$ and $\hat{v}$. Similar to our generator, we concatenate these embeddings into a multimodal embedding vector $\hat{e} \in \mathbb{R}^{T \times \parens{D_{\tilde{l}} + D_{\tilde{v}}}}$. But different from our generator, we pass these multimodal embeddings through a fully-connected classifier network $\textrm{FC}_{\textrm{disc}}$ to obtain class probabilities $c_{\textrm{disc}} \in \bracks{0, 1}$ per sample, as
\begin{equation}
    c_{\textrm{disc}} = \textrm{FC}_{\textrm{disc}}\parens{\hat{e}; \theta_{\textrm{FC}_{\textrm{disc}}}},
    \label{eq:disc_output}
\end{equation}
where $\theta_{\textrm{FC}_{\textrm{disc}}}$ represents the trainable parameters. Our discriminator learns to perform unweighted binary classification between the synthesized face and pose motions and the ground truths in terms of their synchronous affective expressions. Our generator, on the other hand, learns to synthesize samples that our discriminator cannot distinguish from the ground truth based on those affective expressions.

\begin{figure}[t]
    \centering
    \includegraphics[width=\columnwidth]{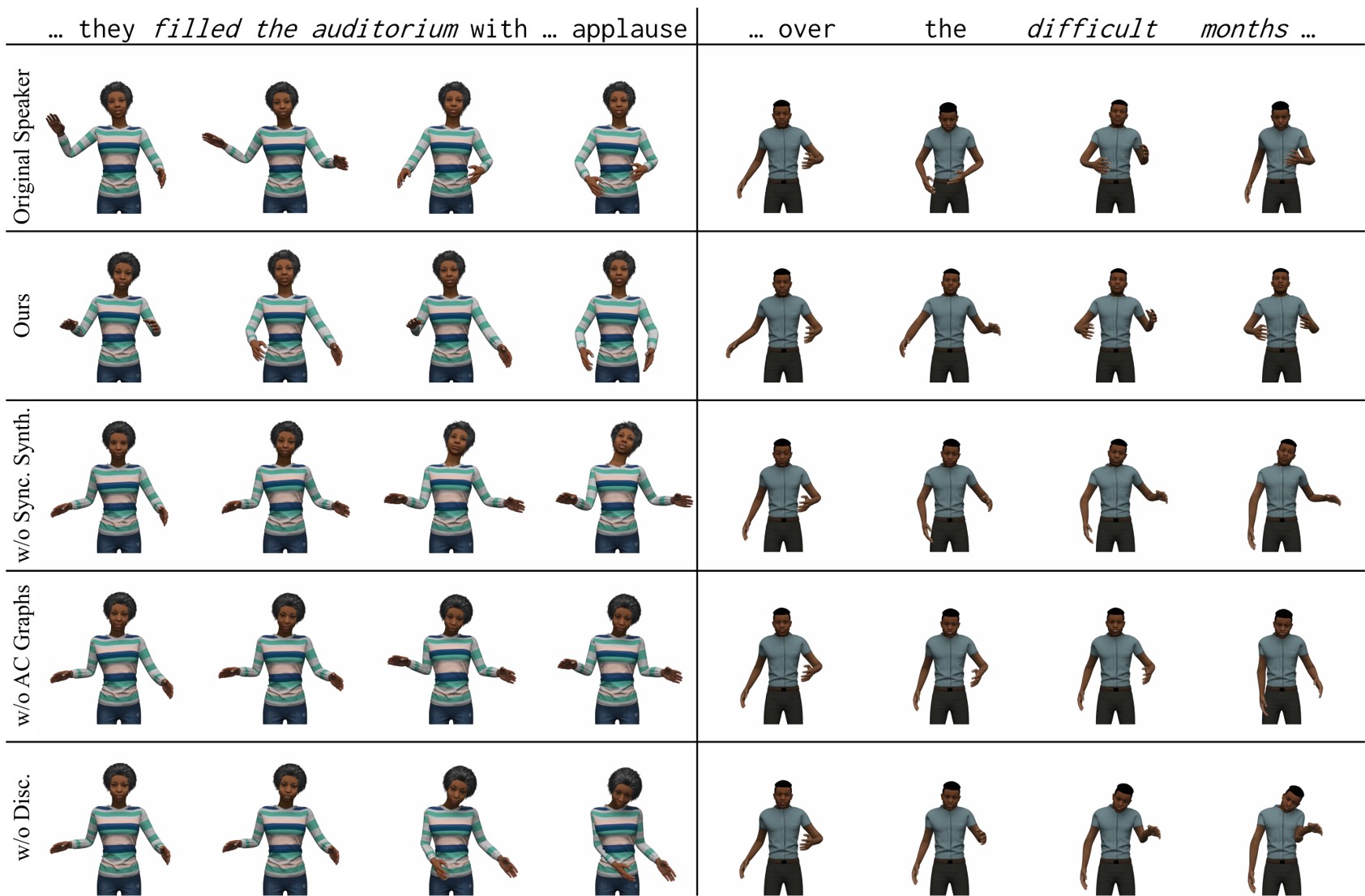}
    \caption{\textbf{Qualitative comparisons.} For the same input speech, represented by the text transcript at the top, we compare the visual quality of our synthesized character motions with the original speaker motions and three of our ablated versions: one without synchronous face and pose synthesis, one without our anatomical component (AC) graphs for faces and poses, and one without our discriminator. We observe that our synthesized motions are visually the closest to the original speaker motions compared to the ablated versions. We elaborate on their visual qualities in Sec.~\ref{subsec:qualitative_comparisons}.}
    \label{fig:qualitative_comparisons}
\end{figure}

\subsubsection{Phoneme Predictor}\label{subsubsec:phoneme_pred}
We train a separate network to learn the positions of the lip landmarks for the different phonemes in the audio. Our synthesis network separately learns the motions of the lip corners denoting the different facial expressions, and we superpose them to the phoneme-based lip shapes to complete the lip motions. Our phoneme predictor predicts the 3D positions of all the landmarks on the inner and the boundaries of the lips over all the $T$ prediction time steps, which we denote as $p_{1:T} \ in \mathbb{R}^{T \times L_{\textrm{lip}} \times 3}$. Following prior approaches~\cite{lipsync}, we design a CNN backbone connected to fully connected blocks to predict the lip landmarks from the spectrograms of the speech inputs. Specifically, given the speech audio waveform $a$, we compute
\begin{equation}
    p_{1:T} = \textrm{PhonemePred}\parens{a; \theta_{\textrm{PhonemePred}}},
\end{equation}
where $\theta_{\textrm{PhonemePred}}$ represents the trainable parameters.

% --------------------------------------------%

\section{TED Gesture+Face Dataset}
We present our TED Gesture+Face Dataset, which we use to train and test our network. We elaborate on collecting and processing our dataset for training and testing.

\paragraph{Dataset Collection.}
The TED Gesture Dataset~\cite{cospeech_gestures} consists of videos of TED talk speakers together with text transcripts of their speeches and their 3D body poses extracted in a global frame of reference. The topics range from personal and professional experiences to discourses on educational topics and instructional and motivational storytelling. The speakers come from a wide variety of social, cultural, and economic backgrounds, and are diverse in age, gender, and physical abilities.

\paragraph{Dataset Processing.}
The 3D poses in the original TED Gesture Dataset~\cite{cospeech_gestures} are view-normalized to face front and center at all time steps. We compute similarly view-normalized 3D face landmarks of the speakers
% (Sec.~\ref{subsec:data_preprocessing})
(Sec.~A.1).
Similar to the original TED dataset, we divide the 3D pose and face landmark sequences into equally-sized chunks of size $T=34$ time steps at a rate of 15 fps. Additionally, to reduce the jitter in the predicted 3D face landmarks and pose joints from each video, we sample a set of ``anchor'' frames at a rate of 5 fps and perform bicubic interpolation to compute the face landmark and pose joint values in the remaining frames. We use the first $4$ time steps of pose and face landmarks as our seed values (Sec.~\ref{subsec:face_pose_synthesis}), and predict the next $30$ time steps. The processed dataset consists of 200,038 training samples, 26,903 validation samples, and 26,245 test samples, following a split of 80\%-10\%-10\%.

% --------------------------------------------%

\section{Training Details}\label{sec:training}
We train our phoneme predictor network using reconstruction losses for the lip shapes. We train our synthesis network using a combination of reconstruction losses for the face and the pose motions, the cross-speaker diversity loss to enforce visual differences in expressions across speakers, and the generative adversarial loss for added regularization. We describe these loss functions and our training and testing procedures.

\subsection{Phoneme Predictor Losses}
We represent our phoneme predictor loss as the robust $\ell_1$-norm reconstruction loss between the ground truth and the synthesized lip landmark positions and velocities over the prediction time steps $T$ as
\begin{equation}
    \mathcal{L}_{\textrm{ph}} = \sum_{t=1}^T{\norm{p^{\parens{\textrm{GT}}}_t - p^{\parens{\textrm{sn}}}_t}_1 + \norm{\Delta_tp^{\parens{\textrm{GT}}}_t - \Delta_tp^{\parens{\textrm{sn}}}_t}_1},
\end{equation}
where the superscripts $\parens{\textrm{GT}}$ and $\parens{\textrm{sn}}$, respectively, denote the ground-truth and the synthesized data. $\Delta_t$ denotes the discrete forward difference between adjacent time steps $t$ and $t-1$.

\subsection{Synchronous Synthesis Network Losses}
We use reconstruction losses to robustly align the outputs of our generator with the corresponding ground-truth face and pose motions. We use the generative adversarial loss to ensure that the synthesized motions are plausible, the affective expressions match the corresponding ground truths, and prevent the mode collapse of only synthesizing singular expressions.
\subsubsection{Reconstruction Losses}\label{subsubsec:rec_loss}
We write our reconstruction losses as the $\ell_1$-norm difference between the ground truth and the synthesized face and pose positions and motions over the $T$ prediction time steps as
\begin{small}
    \begin{align*}
        \mathcal{L}_{\textrm{Rec}} &= \sum_{t=1}^T\norm{\mathcal{F}^{\parens{\textrm{GT}}}_t - \mathcal{F}^{\parens{\textrm{sn}}}_t}_1 + \norm{\mathcal{P}^{\parens{\textrm{GT}}}_t - \mathcal{P}^{\parens{\textrm{sn}}}_t}_1 \\
        &+ \lambda_{\textrm{vel}}\parens{\norm{f^{\parens{\textrm{GT}}}_t - f^{\parens{\textrm{sn}}}_t}_1 + \norm{u^{\parens{\textrm{GT}}}_t - u^{\parens{\textrm{sn}}}_t}_1} \\
        &+ \lambda_{\textrm{acc}}\parens{\norm{\Delta_tf^{\parens{\textrm{GT}}}_t - \Delta_tf^{\parens{\textrm{sn}}}_t}_1 + \norm{\Delta_tu^{\parens{\textrm{GT}}}_t - \Delta_tu^{\parens{\textrm{sn}}}_t}_1}, \numberthis \label{eq:synth_rec_loss}
    \end{align*}
\end{small}
where $\lambda_{\textrm{vel}}$ and $\lambda_{\textrm{acc}}$ are the relative weighting factors. We use the velocity and acceleration losses to enforce smoothness in the synthesized motions by reducing jitters.

\subsubsection{Cross-Speaker Diversity Loss}
Our cross-speaker diversity loss $\mathcal{L}_{\textrm{CSD}}$ follows that of Yoon et al.~\cite{trimodal}, consisting of a ranking loss between the ground-truth face and pose motions, and the synthesized face and pose motions using the same speaker as the ground-truth (positive example) and a randomly chosen different speaker (negative example).

\subsubsection{Generative Adversarial Loss}\label{subsubsec:adv_loss}
The generative adversarial loss consists of opposing losses $\mathcal{L}_{\textrm{Gen}}$ for the generator and $\mathcal{L}_{\textrm{Dis}}$ for the discriminator, following a min-max optimization strategy~\cite{gan}. We write these losses as
\begin{align}
    \mathcal{L}_{Gen} &= -\mathbb{E}\bracks{\log\parens{c_{\textrm{disc}}^{\textrm{GT}}}},\\
    \mathcal{L}_{Dis} &= -\mathbb{E}\bracks{\log\parens{c_{\textrm{disc}}^{\textrm{GT}}}} - \mathbb{E}\bracks{\log\parens{1 - c_{\textrm{disc}}^{\textrm{sn}}}},
    \label{eq:adversarial_loss}
\end{align}
where $c_{\textrm{disc}}$ denotes the output of our discriminator network (Eq.~15). This loss adds plausibility to our synthesized samples by forcing them to have affective expressions similar to those of the corresponding ground-truth samples.

\subsection{Training Procedure}
We train our phoneme predictor network using the Adam optimizer~\cite{adam} with $\beta_1 = 0.5$, $\beta_2 = 0.999$, a batch
size of $1024$, and a learning rate of $10^{-3}$ for 500 epochs. We train our synthesis network using the Adam optimizer~\cite{adam} with $\beta_1 = 0.5$, $\beta_2 = 0.999$, a batch
size of $256$, and learning rates of $10^{-4}$ for our generator and $5 \times 10 ^{-5}$ for our discriminator, both decayed by a factor of $0.999$ per epoch, for 1000 epochs. We train both our phoneme detector network and our synthesis network on an NVIDIA GeForce RTX 2080 Ti GPU, which takes $3$ seconds and $7$ seconds per epoch, respectively.

\subsection{Testing and Rendering}
We provide the details of the testing procedure of our network and the rendering of our synthesized outputs in a 3D environment.

\subsection{Testing Procedure and Mapping to Digital Characters}
Each test sample for our network consists of a speech audio waveform, the corresponding text transcript, a speaker ID, and the speaker's seed face and pose motions. Our phoneme predictor network provides the lip sync for the given speech audio, and the generator of our synthesis network provides the required face and pose motions. We superpose the lip landmarks given by our phoneme predictor network with the lip corner landmarks given by our generator at each prediction time step to obtain the complete lip motions of the speaker. We map these motions to a rigged 3D human upper-body mesh in Blender. For mapping the face motions, we set a one-to-one mapping between our face landmarks and the landmarks on the face of the human mesh, and use them as control points for the facial motions of the mesh. For mapping the pose motions, we use FABRIK~\cite{fabrik} to obtain the joint rotations given our predicted joint positions and use those rotations to animate the rigged human mesh.

\subsection{Rendering and Visualization}
Given an input speech audio, we can synthesize the motions for our pre-rigged digital characters at an interactive rate of about 250 frames per second on an NVIDIA GeForce RTX 2080 Ti GPU. We design our digital environment using Blender. For each of our digital characters, we place them on a stage and position the camera such that it looks front and center at the agent. As the character narrates the input speech audio using our synthesized face and upper-body expressions, we slowly pan the camera in to get a more focused view of those expressions. Since we do not synthesize any lower-body motions, our digital characters stay standing at their initial positions during the entire narration. The full video demos are available with our supplementary material.

% --------------------------------------------%

\section{Experiments and Results}
We run quantitative experiments using ablated versions of our method as baselines. We note that \citeauthor{speech_driven_conv_gestures}~\cite{speech_driven_conv_gestures} retrain their network separately for individual speakers belonging to the same profession (talk show hosts), making it unsuitable for our generalized paradigm consisting of less than 50 samples each of multiple, diverse speakers. \citeauthor{talkshow}~\cite{talkshow} use VQ with transformers to synthesize faces and gestures, but are limited to the same set of fixed speakers. We also conducted a web-based user study to evaluate the qualitative performance of our method.

\begin{table}[t]
    \centering
    \caption{\textbf{Quantitative evaluations.} Comparison with existing co-speech gesture synthesis methods and our ablated versions (Sec.~\ref{subsec:baselines}) on the metrics MALE (in mm), MAJE (in mm), MAcE for landmarks (MAcE-LM) (in mm/s$^2$), MAcE for poses (MAcE-P) (in mm/s$^2$), FLD, and FGD (Sec.~\ref{subsec:eval_metrics}). Lower values are better, bold indicates \textbf{best}, and underline indicates \underline{second-best}.}
    \label{tab:quant_eval}
    \resizebox{\columnwidth}{!}{
    \begin{tabular}{lrrrrrr}
        \toprule
        Method & MALE & MAJE & MAcE-LM & MAcE-P & FLD & FGD \\
        \midrule
        Seq2Seq~\cite{cospeech_gestures} & -- & 45.62 & -- & 6.33 & -- & 6.62 \\
        S2G-IS~\cite{individual_gesture_styles} & -- & 45.11 & -- & 7.22 & -- & 6.73 \\
        JEM~\cite{language2pose} & -- & 48.56 & -- & 4.31 & -- & 5.88 \\
        GTC~\cite{trimodal} & -- & 27.30 & -- & 3.20 & -- & 4.49 \\
        Speech2AffectiveGestures~\cite{s2ag} & -- & 24.49 & -- & 2.93 & -- & 3.54 \\
        SpeechGestureMatching~\cite{motion_matching} & -- & \underline{21.10} & -- & \underline{2.75} & -- & \underline{2.64} \\
        \midrule
        Ours w/o Face Synthesis & -- & 28.32 & -- & 3.89 & -- & 4.01 \\
        Ours w/o Pose Synthesis & 11.76 & -- & 9.38 & -- & 22.65 & -- \\
        Ours w/o Vel.+Acc. Losses & 26.33 & 24.41 & 21.69 & 7.58 & 27.54 & 7.72 \\
        Ours w/o Discriminator & 14.62 & 27.40 & 13.44 & 11.60 & 31.93 & 8.79 \\
        Ours w/o Face AC Graph & 13.05 & 25.97 & 14.24 & 2.74 & 25.61 & 2.25 \\
        Ours w/o Pose AC Graph & 11.84 & 25.46 & 8.12 & 13.88 & 19.23 & 6.94 \\
        Ours w/o Synchronous Synthesis & \underline{10.72} & 25.03 & \underline{7.83} & 3.22 & \underline{18.03} & 3.92 \\
        \textbf{Ours} & \textbf{9.00} & \textbf{18.36} & \textbf{6.34} & \textbf{2.52} & \textbf{15.02} & \textbf{1.79} \\
        \bottomrule
    \end{tabular}
    }
\end{table}

\subsection{Baselines}\label{subsec:baselines}
We use seven ablated versions of our method as baselines. The first two ablations correspondingly remove the entire face (Figs.~\ref{fig:face_encoder}, \ref{fig:face_decoder}) and pose components (Figs.~\ref{fig:pose_encoder}, \ref{fig:pose_decoder}) from our network, making our network learn only talking head and only co-speech gesture syntheses. The third ablation removes the velocity and acceleration losses from our reconstruction loss
% (Eqn.~\ref{S-eq:synth_rec_loss})
(Eqn.~C.2)
, leading to jittery motions. The fourth ablation removes the discriminator and its associated losses
% (Eqn.~\ref{eq:S-adversarial_loss})
(Eqn.~C.4)
from our training pipeline, leading to unstable motions without appreciable expressions. The fifth and the sixth ablations correspondingly remove the ``higher-level'' anatomical component (AC) graphs of the faces (Eqn.~\ref{eq:stgcn_face_2}) and the poses (Eqn.~\ref{eq:stgcn_pose_2}), leading to reduced movements. The final ablation trains the face and the pose expressions separately, learning marginal embeddings for the two modalities based on the speech but not attending to their mutual synchronization. This ablation directly evaluates the co-speech motions when combining separately synthesized face and pose expressions. For completeness, we also compare with co-speech gesture synthesis methods that only synthesize body poses. We evaluate all the methods on our TED Gesture+Face Dataset.

\subsection{Evaluation Metrics}\label{subsec:eval_metrics}
Inspired by prior work~\cite{trimodal}, we evaluate using four \textit{reconstruction errors} and two \textit{plausibility errors} (PEs). Our reconstruction errors include the mean absolute landmark error (MALE) for the faces, the mean absolute joint error (MAJE) for the poses, and their respective mean acceleration errors (MAcEs). MALE and MAJE indicate the overall fidelity of the synthesized samples w.r.t. the corresponding ground truths, and the MAcEs indicate whether or not the synthesized landmarks and poses have regressed to their mean absolute positions. To report these metrics, we multiply our ground truth and synthesized samples by a constant scaling factor such that they all lie inside a bounding box of diagonal length 1 m. For our PE, we use the Fr\'echet Gesture Distance (FGD) designed by~\cite{trimodal} to indicate the perceived plausibility of the synthesized poses. To similarly indicate the perceived plausibility of the synthesized face landmarks, we also design the Fr\'echet Landmark Distance (FLD). We train an autoencoder network to reconstruct the set of face landmarks at all time steps for all the samples in the training set of our TED Gesture+Face Dataset. To compute FLD, we then obtain the Fr\'echet Inception Distance~\cite{fid} between the encoded features of the ground truth and the synthesized samples.

\subsection{Quantitative Evaluations}
We show our quantitative evaluations in Table~\ref{tab:quant_eval}.

\paragraph*{Comparison with Co-Speech Gesture Synthesis.} Since co-speech gesture synthesis methods do not synthesize face expressions, we leave those numbers blank. For these methods, we have taken the numbers reported by \citeauthor{s2ag}~\cite{s2ag}. For the method of SpeechGestureMatching~\cite{motion_matching}, we retrain their method on the TED Gesture Dataset to report the numbers. However, we were unable to perform similar comparative evaluations with co-speech face synthesis methods as existing methods synthesize dense landmarks~\cite{audio_driven_face} or blendshape-like features~\cite{voca}, which cannot be mapped one-to-one with our sparser face landmarks.

\paragraph*{Comparison with Ablated Versions}
Removing either the face or the gesture components of our network leads to poorer values across the board than using both. Without the velocity and acceleration losses, the motions are jittery, and the MAcE losses are higher, especially MAcE for the face landmarks. Without the discriminator, the synthesized samples suffer from mode collapse and often produce implausible motions, leading to higher values across the board. Without the AC graphs, there are fewer movements in the synthesized motions and the reconstruction errors are higher. When synthesizing face and pose expressions separately and not synchronizing them, we observe some mismatches in when the expressions from either modality appear and how intense they are. This indicates that synchronous synthesis of facial expressions and body gestures leads to more accurate and plausible movements for both the modalities, including a $30\%$ improvement on MALE and a $21\%$ improvement on MAJE, compared to trivially combining synthesized outputs of the individual modalities.

\subsection{Qualitative Comparisons}\label{subsec:qualitative_comparisons}
We visualize some of our synthesized samples in Fig.~\ref{fig:qualitative_results} and provide more results in our supplementary video. We observe the synchronization between the face and the pose expressions for two contrasting emotions. We also visually compare with the original speaker motions rendered using their face landmarks and the poses extracted from the videos and three of our ablated versions in Fig.~\ref{fig:qualitative_comparisons}. The original speaker motions provide an ``upper bound'' of our performance. The three ablated versions we compare with are: one without the synchronous synthesis, one without our face and pose AC graphs, and one without our discriminator. The ablated versions without either the face or the pose synthesis, without the velocity and acceleration losses, and without our discriminator are visually inferior in obvious ways. Therefore, we leave them out. Without either face or pose synthesis, that modality remains static while there is movement in the other. Without the velocity and the acceleration losses, the overall motions regress to the mean pose. Without our discriminator, our generator often fails to understand plausible movement patterns, leading to unnatural limb and body shapes. Of these, we only keep the ablations without our discriminator as our ``lower bound'' baseline because, unlike the other two, this ablation has visible movements in both the face and the pose modalities.

\section{User Study}\label{sec:user_study}
We conducted a web-based user study to evaluate the visual quality of our synthesized motions in terms of their plausibility and synchronization.

\begin{figure*}[t]
    \centering
    \begin{subfigure}[b]{0.48\textwidth}
        \centering
        \includegraphics[width=\textwidth]{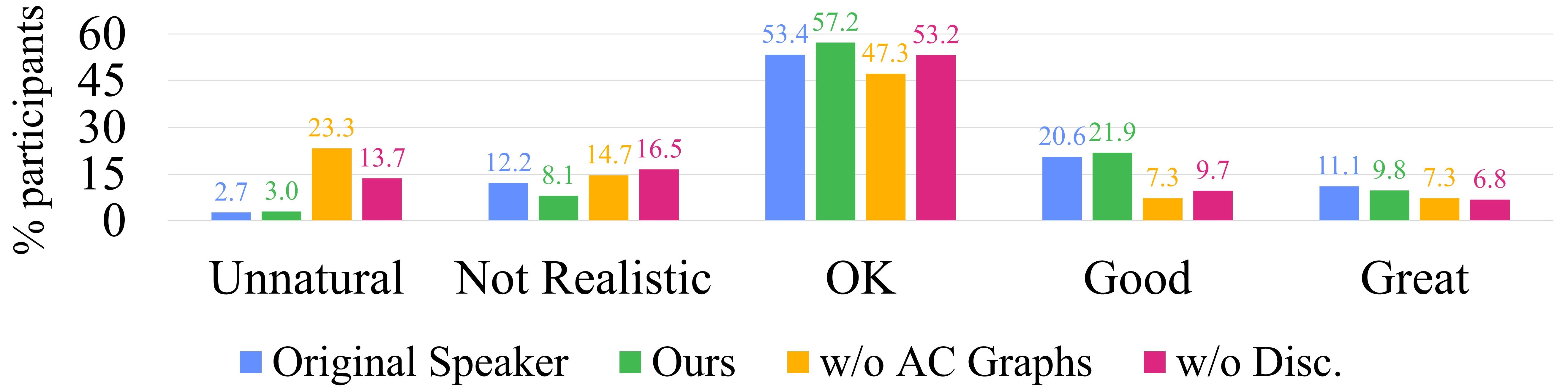}
        \caption{\textbf{Set 1: Motion plausibility.} Compared to the ablated versions, we observe a higher distribution of ``OK'' or better for the motions of the original speakers and our synthesized agents. The modes of all the distributions are on ``OK'', implying that the corresponding participants found the visual qualities of all the motions to be reasonable.}
        \label{fig:plausibility_s1}
    \end{subfigure}
    \hfill
    \begin{subfigure}[b]{0.48\textwidth}
        \centering
        \includegraphics[width=\textwidth]{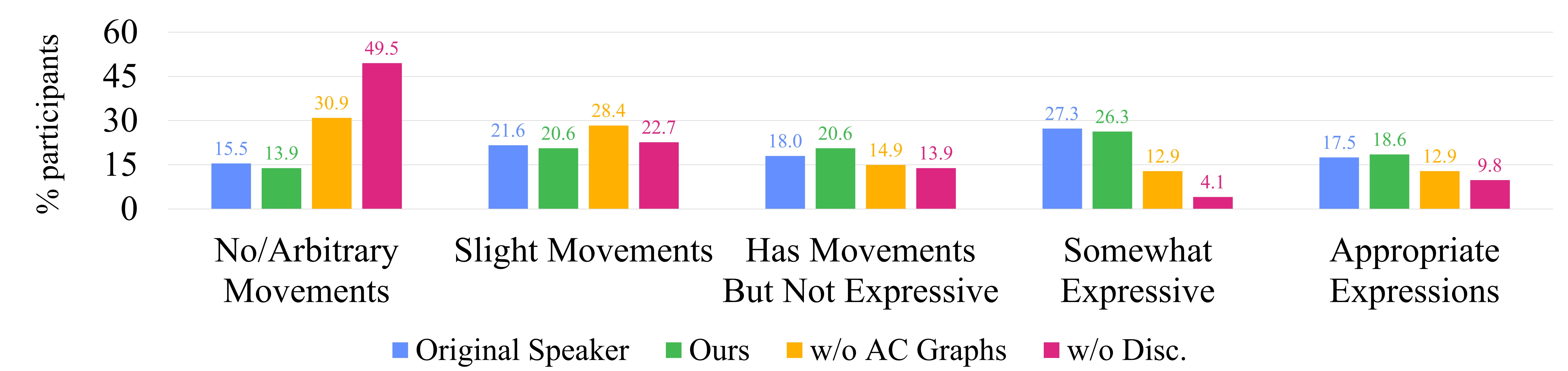}
        \caption{\textbf{Set 1: Synchronization between the face and the pose expressions given the speech.} Compared to the ablated versions, we obverse clear preferences for the motions of the original speakers and our synthesized agents. The modes of the distributions for these two types of motions are on ``somewhat expressive'' while the modes of the two ablated versions are on ``no/arbitrary movements''.}
        \label{fig:sync_s1}
    \end{subfigure}
    ~
    \begin{subfigure}[b]{0.48\textwidth}
        \centering
        \includegraphics[width=\textwidth]{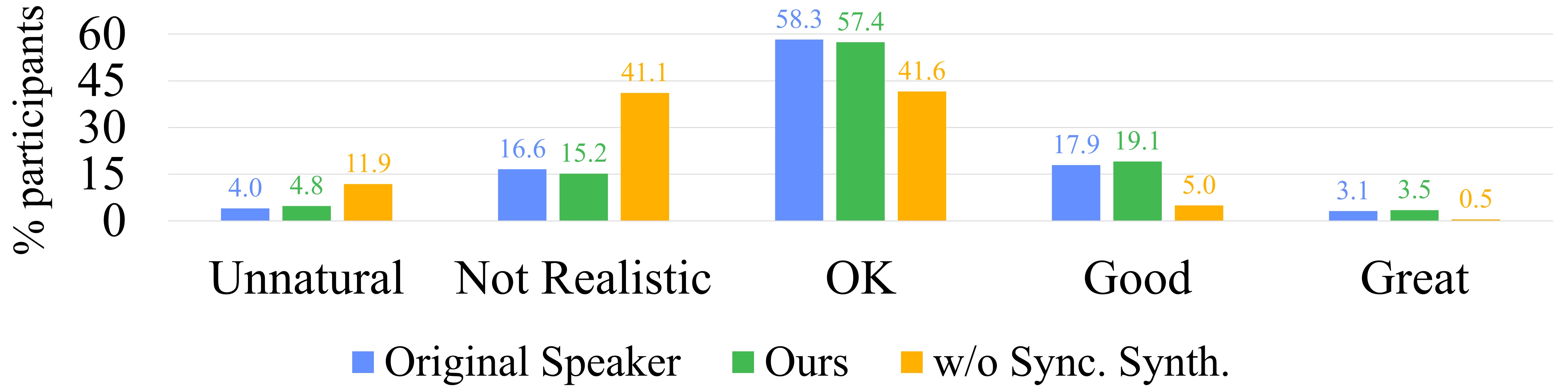}
        \caption{\textbf{Set 2: Motion plausibility.} Compared to the ablated version without synchronous synthesis, we observe a higher distribution of ``OK'' or better for the motions of the original speakers and our synthesized agents. Similar to the motion plausibility in set 1, we observe modes of all the distributions on ``OK''.}
        \label{fig:plausibility_s2}
    \end{subfigure}
    \hfill
    \begin{subfigure}[b]{0.48\textwidth}
        \centering
        \includegraphics[width=\textwidth]{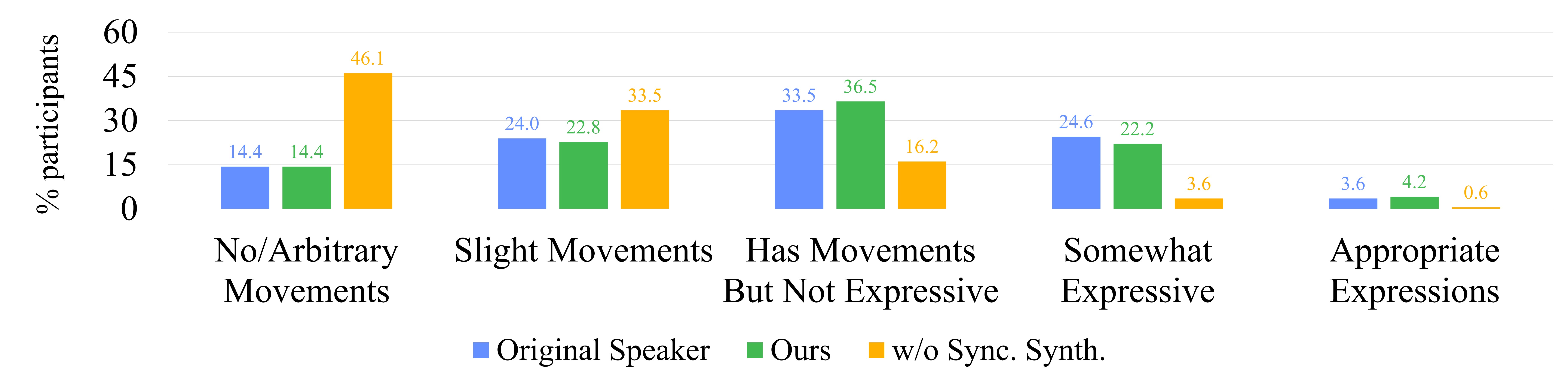}
        \caption{\textbf{Set 2: Synchronization between the face and the pose expressions given the speech.} We again obverse clear preferences for the motions of the original speakers and our synthesized agents compared to the ablated version without synchronous synthesis. However, in contrast to the same study in set 1, we notice the modes of the distributions for the first two types of motions are one point lower on the Likert scale, whereas the mode for the ablated version remains on ``no/arbitrary movements''. We hypothesize this to be the consequence of removing the other ablated versions from the participants' cognitive window: in the absence of other variants, the participants focused more closely on the relative qualities of asynchronous vs. synchronous motions and assessed them more critically.}
        \label{fig:sync_s2}
    \end{subfigure}
    \caption{\textbf{Distributions of the user study responses.} Likert-scale response distributions to the two sets of motions rendered using the five different types of face landmark and pose data (Sec.~\ref{sec:user_study}). We show the distributions of each of the five Likert-scale points for each type of motion as a percentage of the total responses across all the groups in each set.}
    \label{fig:user_study}
\end{figure*}

\paragraph{Setup.}
A total of 90 participants participated in our user study. All participants were aged 18 years or older, and had normal or corrected to normal vision and hearing. Each participant observed two sets of character motions. There were eight groups of motions in each set, each group having a unique input speech. In the first set, there were four types of motions in each group corresponding to the same speech: the original speaker motions rendered using their face landmarks and the poses extracted from the video, and motions rendered using the face landmarks and poses synthesized by our network and two of its ablated versions. One ablated version was without using the face and pose anatomical component (AC) graphs for training, and one without our discriminator. In the second set, there were three types of motions in each group corresponding to the same speech: the original speaker motions, motions rendered using the face landmarks and poses synthesized by our network, and the ablated version using asynchronously synthesized faces and poses. Our motivation to separately compare with the asynchronously synthesized motions was to eliminate distractors from other motions and enable our participants to focus more closely on the synchronization between the face and the pose expressions. We randomized the order of these motions in each group in each set and kept the order unknown to the participants. We did not present our other ablated versions to the participants as they did not have sufficient motion and were visually inferior in obvious ways.

\begin{table}[t]
    \centering
    \caption{\textbf{Likert-scale score statistics.} We compute the mean and the standard deviation of the Likert-scale scores across all the motions. For the mean scores, higher values are better, bold indicates \textbf{best}, and underline indicates \underline{second-best}.}
    \label{tab:user_study_mean_scores}
    \resizebox{\columnwidth}{!}{
    \begin{tabular}{llrrrr}
        \toprule
        & Synthesis type & \multicolumn{2}{c}{Plausibility} & \multicolumn{2}{c}{Synchronization} \\
        \cmidrule{3-6}
        && Mean & St. Dev. & Mean & St. Dev. \\
        \midrule
        \multirow{4}{*}{Set 1} & Original Speaker & \underline{3.25} & 0.90 & \underline{3.10} & 1.34 \\
        & Ours & \textbf{3.27} & 0.86 & \textbf{3.15} & 1.32 \\
        & w/o AC Graphs & 2.61 & 1.14 & 2.48 & 1.38 \\
        & w/o Disc. & 2.79 & 1.02 & 2.02 & 1.30 \\
        \midrule
        \multirow{3}{*}{Set 2} & Original Speaker & \underline{2.99} & 0.80 & \textbf{2.79} & 1.08 \\
        & Ours & \textbf{3.01} & 0.82 & \textbf{2.79} & 1.07 \\
        & w/o Synchronous Synthesis & 2.41 & 0.78 & 1.79 & 0.88 \\
        \bottomrule
    \end{tabular}
    }
\end{table}

\paragraph{Evaluation Process.}
Our aim in the user study is to evaluate our synthesized motions on two key aspects: \textit{(i)} how plausible they appear to human observers compared to the motions of the original speakers and the ablated versions, and \textit{(ii)} whether synchronous synthesis of face and pose expressions produces perceptible improvements over asynchronous synthesis. To evaluate plausibility, we ask the participants to rate each motion in each group in each set on ``how natural the motion looks'' on a five-point Likert scale, with the options ``very unnatural'' (worst), ``not realistic'', ``looks OK'' (average), ``looks good'', and ``looks great'' (best). To evaluate the effect of synchronous synthesis, we ask the participants to observe the face and the pose movements in each motion in each group in each set and rate them on ``how the face and the pose sync with the speech'' on a five-point Likert scale, with the options ``no/arbitrary movements'' (worst), ``slight movements'', ``has movements, but are not expressive'' (average), ``somewhat expressive movements'', and ``have movements with appropriate expressions'' (best).

\begin{figure}[t]
    \centering
    \includegraphics[width=\columnwidth]{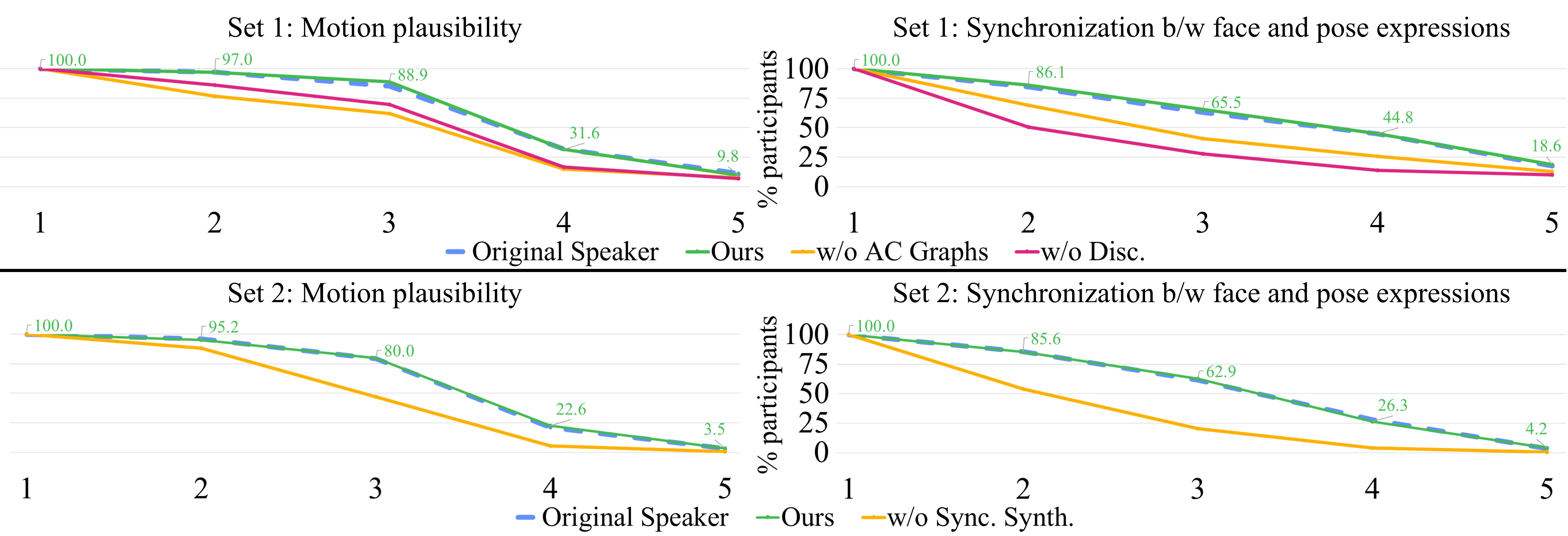}
    \caption{\textbf{Cumulative lower-bound of participant responses.} We plot the cumulative lower-bound (LB) percentage of responses across the Likert-scale scores for each type of character motion in each set. A cumulative LB percentage $X$ for a Likert-scale score $s$ denotes $X\%$ of responses had a score of $s$ or higher. We observe that the curve for our synchronously synthesized motions stays at the top, indicating that the participants preferred it over the other motions.}
    \label{fig:cumulative_lb_plots}
\end{figure}

\paragraph{Results.}
Since we randomly select the speech for each of the eight groups of motions each participant watched, and we also randomized the order of the motions in each group in each set, we can consider the participants' responses in each group to be independent of all the other groups. Thus, we aggregate their responses to each type of motion across all the groups within a set to obtain the overall distributions of the Likert-scale scores of the motions for that set. We show these distributions for each of the two questions on plausibility and synchronization in each set in Fig.~\ref{fig:user_study}. We also report the Likert-scale score statistics for each type of motion on the two questions in each set in Table~\ref{tab:user_study_mean_scores}. For the purpose of scoring, we assign scores 1 through 5, with 1 for ``worst'' and 5 for ``best''. We observe that the scores for our synthesized samples are comparable to the corresponding original speaker motions and significantly better than the ablated versions. To further affirm this, we plot the cumulative lower bound of participant responses for each Likert-scale score for each type of motion in each set in Fig.~\ref{fig:cumulative_lb_plots}. We note that the scores for our synchronously synthesized samples remain close to the original speaker scores and consistently above the other ablated versions, indicating a clear preference. Overall, in the two sets, $88.89\%$ and $80.00\%$ participants respectively marked our synchronously synthesized motions 3 or above on the first question, and $65.46\%$ and $62.87\%$ participants respectively marked 3 or above on the second question. This indicates that the majority of participants found the motions satisfactory.

% --------------------------------------------%

\section{Conclusion, Limitations and Future Work}
We have presented a method to synthesize synchronous co-speech face and pose expressions for 3D digital characters. Our method learns to synthesize these expressions from 3D face landmarks and 3D upper-body pose joints computed directly from videos. Our work also has some limitations. We use sparse face landmarks and pose joints to synthesize co-speech face and pose expressions. To synthesize more fine-grained expressions, we plan to extract more detailed face meshes and additional pose joints from videos. Further, given the sparsity of our face and pose representations and the noise associated with extracting them from videos, the quality of our synthesized motions does not match those synthesized from high-end facial scans and motion-capture data, and using parameter-dense, compute-heavy methods, such as denoising diffusion. We aim to bridge this gap by building techniques to develop more robust face and pose representations from videos.
We also plan to combine our work with lower-body actions such as sitting, standing, and walking to synthesize 3D-animated digital humans in a wider variety of scenarios. In terms of its running-time cost, our method uses commercial GPUs to obtain real-time performance. We plan to explore knowledge distillation techniques to reduce our running-time cost and implement our method in real-time on commodity devices such as digital personal assistants.

% --------------------------------------------%

\section*{Acknowledgments}
This work was partly supported by ARO Grants W911NF2110026, W911NF2310046, W911NF2310352, and Army Cooperative Agreement W911NF2120076.

% --------------------------------------------%

{
    \small
    \bibliographystyle{ieeenat_fullname}
    \bibliography{main}
}


\end{document}